\documentclass{article}
\usepackage{spconf,amsmath,epsfig}

\let\OLDthebibliography\thebibliography
\renewcommand\thebibliography[1]{
  \OLDthebibliography{#1}
  \setlength{\parskip}{0pt}
  \setlength{\itemsep}{0pt plus 0.3ex}
}

\title{USING DEEP LEARNING TO COUNT ALBATROSSES FROM SPACE}
\name{Ellen Bowler\textsuperscript{1*}, Peter T. Fretwell\textsuperscript{2}, Geoffrey French\textsuperscript{1}, Michal Mackiewicz\textsuperscript{1}  \thanks{* Corresponding author. Email: e.bowler@uea.ac.uk}}
\address{\textsuperscript{1}School of Computing Sciences, University of East Anglia, Norwich, UK \\
\textsuperscript{2}British Antarctic Survey, Maddingley Road, Cambridge, UK}
\begin{document}
%

\onecolumn

\vspace*{\fill}
\begin{center}
Copyright 2019 IEEE. Published in the IEEE 2019 International Geoscience \& Remote Sensing Symposium (IGARSS 2019), scheduled for July 28 - August 2, 2019 in Yokohama, Japan. Personal use of this material is permitted. However, permission to reprint/republish this material for advertising or promotional purposes or for creating new collective works for resale or redistribution to servers or lists, or to reuse any copyrighted component of this work in other works, must be obtained from the IEEE. Contact: Manager, Copyrights and Permissions / IEEE Service Center / 445 Hoes Lane / P.O. Box 1331 / Piscataway, NJ 08855-1331, USA. Telephone: + Intl. 908-562-3966.
\end{center}
\vspace*{\fill}

\twocolumn

\maketitle
\begin{abstract}
 
In this paper we test the use of a deep learning approach to automatically count Wandering Albatrosses in Very High Resolution (VHR) satellite imagery. We use a dataset of manually labelled imagery provided by the British Antarctic Survey to train and develop our methods. We employ a U-Net architecture, designed for image segmentation, to simultaneously classify and localise potential albatrosses. We aid training with the use of the Focal Loss criterion, to deal with extreme class imbalance in the dataset. Initial results achieve peak precision and recall values of approximately 80\%. Finally we assess the model's performance in relation to inter-observer variation, by comparing errors against an image labelled by multiple observers. We conclude model accuracy falls within the range of human counters. We hope that the methods will streamline the analysis of VHR satellite images, enabling more frequent monitoring of a species which is of high conservation concern. 

\end{abstract}
\section{Introduction}
\label{sec:intro}

Albatrosses have been in decline in recent decades, and now encompass the highest proportion of threatened species of any bird family \cite{phillips2016conservation}. However, given many colonies exist on remote islands in the southern oceans, conducting ground or aerial surveys can be challenging and accurate population data is limited. Recently the advent of Very High Resolution (VHR) satellite imagery has provided an alternative - to count certain species of large 'Great Albatrosses' directly from space. For example in 2017 \cite{fretwell2017using} used 31-cm resolution WorldView-3 (WV-3) imagery to count Wandering and Northern Royal Albatross populations in South Georgia and the Chatham Islands.

While VHR satellites have the potential to drastically improve the spatial coverage and frequency of wildlife surveys, manually analysing imagery is time consuming, tedious and expensive. Analysts are faced with the task of identifying four to five pixel white dots in imagery covering many kilometres, resulting in counts which can vary significantly between observers (\textit{inter-observer variation}). This strongly motivates the development of automated approaches, which could solve the problem by giving quick, consistent and comparable results with known associated errors.   

In this paper we test the application of a Deep Learning approach to the task of detecting Wandering Albatrosses \textit{Diomedea exulans} in WV-3 imagery. We use a manually labelled dataset of approximately 2000 birds to test and develop our methods. In comparison to more standard image datasets (e.g COCO, PASCAL VOC) there are several factors we consider with this specific task. 

Firstly, as images are collected by the same satellite, all albatrosses appear at a fixed scale and are not subject to occlusion or view point changes. In this respect many of the hardest challenges faced by object detection methods are removed. With this in mind we implement a comparatively simple U-net architecture \cite{ronneberger2015u}, designed for semantic segmentation, to the task. This allows for simultaneous classification and localization of albatrosses in a single stage, outputting a heatmap of detections which matches the dimensions of the input image. Additionally, as the number of albatross pixels in the imagery is vastly outweighed by background instances, we give extra consideration to addressing class imbalance. We test the effectiveness of the Focal Loss \cite{lin2017focal} as the optimization criterion, which gives extra weighting to sparse, hard to classify examples.

Finally, while most CNN based detection methods are trained on clearly discernible and reliably ground truthed images, in this problem we rely solely on ground truth labels obtained by manual analysis of the satellite imagery. As noted, these labels are subject to inter-observer variation, introducing uncertainty into the methods. The success of any automated approach must be assessed with this in mind. We therefore test our final approach on an image labelled by eight different observers, to gain an insight into the scale of this variation. 

\begin{figure*}[!t]
\centering
\includegraphics[width=\textwidth]{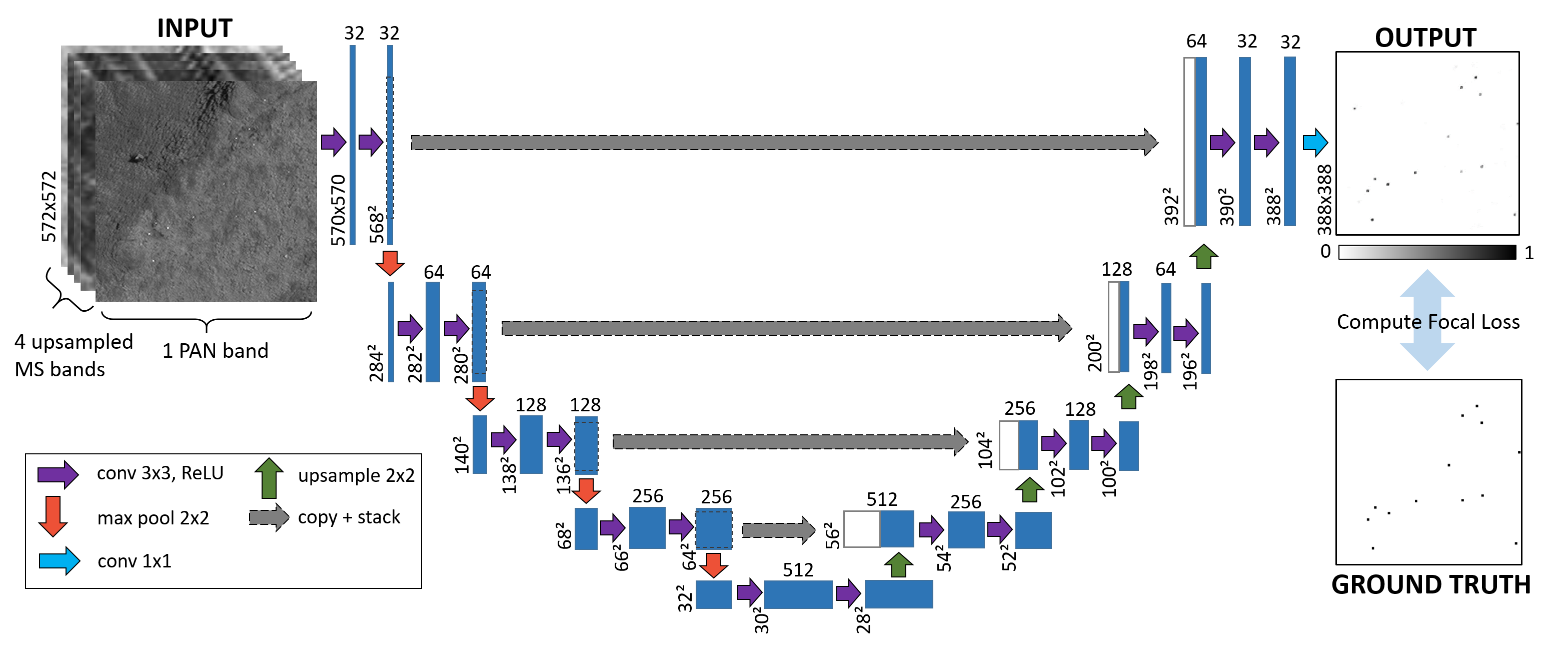}
\caption{U-net architecture, showing image dimensions and the number of feature maps after each operation.}
\label{fig:unet}
\end{figure*}

\section{DATASET DESCRIPTION}
\label{sec:dataset}
Our dataset consists of WV-3 images of four separate nesting colonies of Wandering Albatrosses. Colonies are located on Bird and Annenkov Island in South Georgia, Ile des Apotres in the Crozet Islands, and Grand Coulee on the west coast of Kerguelen Island (detailed descriptions in \cite{fretwell2017using} and \cite{weimerskirch2018status}). Images were captured at 31-cm resolution in the panchromatic (PAN) band, and at 1.24-m resolution in four multispectral (MS) bands (blue, green, red and near-IR1).

A full description of the manual labelling procedure is described in \cite{fretwell2017using}. Briefly, MS imagery was pansharpened to generate a 31-cm resolution colour image, in which Wandering Albatrosses appear as four to five pixel white dots against their green nesting habitat. These dots were counted by the same observer for all images, and were manually digitized using ArcMap 10.1. The number of albatrosses per image differed (Bird Island - 935, Annenkov - 161, Apotres - 171 and Grand Coulee - 649), with a total of 1916. In addition counts from seven different observers were obtained for the Bird Island colony using the same procedure. Count values varied (508, 613, 683, 862, 868, 720 and 1016), as did observers' experience. This included those who had no prior knowledge of the study site or species, researchers highly familiar with the colony, and satellite remote sensing experts.

\section{NETWORK}
\label{sec:Network}

\subsection{Network architecture}
\label{ssec:architecture}
The U-net architecture used in our experiments is presented in Fig \ref{fig:unet}, and follows a similar structure to that described in the original paper \cite{ronneberger2015u}. The contracting path (left) follows the typical architecture of a CNN, applying repeated $3\times3$ convolutions, ReLU activations, and $2\times2$ max pooling to extract features from input images. The expanding path (right) upsamples feature maps and concatenates them with higher resolution information cropped and copied from the corresponding layer in the contracting path. This allows for precise localization of classified pixels. To minimise information loss at the edge of images, we choose not to use padding in convolution operations, thus the output predictions are of reduced size ($388\times388$ compared to $572\times572$ inputs). Experiments also showed no performance gain when using learnt upsampling (through upconvolution), so we favour bilinear upsampling for simplicity. 

\subsection{Training}
\label{sssec:training}
In order to train and test U-net, we cropped $572\times572$ square patches from each satellite image. The MS bands were upsampled using bilinear interpolation to match the dimensions of the PAN image (we note in the future alternative methods, such as pansharpening algorithms, could be tested). Binary segmentation maps (of size $388\times388$) showing albatross locations as $3\times3$ square points were generated from ArcMap shape files. We selected 400 patches from each island (including all containing Albatrosses), and reserved 25\% for testing. This resulted in a training and test set of 1200 and 400 patches respectively, with each island represented proportionally. 

To train the model we minimize the Focal Loss, proposed by \cite{lin2017focal} as a method for addressing extreme foreground-background class imbalance. It works by adding a simple modulating factor to the standard cross entropy criterion, which places more focus on hard, misclassified examples. If $y \in \lbrace\pm1\rbrace$ denotes the ground truth class and $p \in [0, 1]$ is the model's estimated probability for the class with label $y=1$, then the focal loss can be expressed as:
\begin{equation}
FL(p_{t}) = -(1 - p_{t})^{\gamma}\log(p_{t}) 
\end{equation}

\begin{equation*}
\text{where}\quad p_{t} =\begin{cases}
    p, & \text{if $y=1$}.\\
    1-p, & \text{otherwise}.
  \end{cases}
\end{equation*}
Increasing the \textit{focusing parameter} $\gamma \geq 0$ reduces the loss contribution from easy to classify background examples. We ran experiments to assess the best choice for $\gamma$, and trained the model using the Adam optimizer, a learning rate of 0.0001 and a mini-batch size of 4.

In our second set of experiments we additionally test the network over the whole Bird Island dataset (approximately 1100 patches), and assess results in comparison to multiple observers' labels. To remove bias we retrain the network using only patches from the three other islands (300 patches from each). This also allows us to see how well the network generalizes to completely unseen islands and colonies. All other training parameters are kept the same as previously described.

\begin{figure}[b!]
\centering
\includegraphics[width=\linewidth]{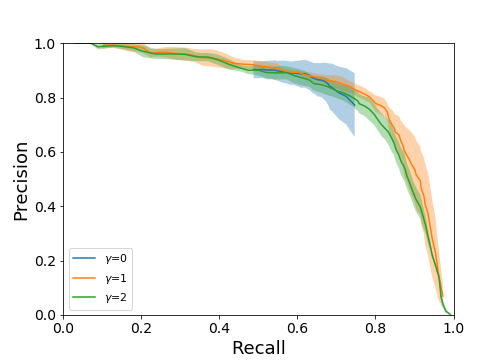}
\caption{Averaged precision-recall curves for different choices of $\gamma$, with error bars showing standard deviation.}
\label{fig:Gamma}
\end{figure}	

\begin{figure}[t]
\centering
\includegraphics[width=\linewidth]{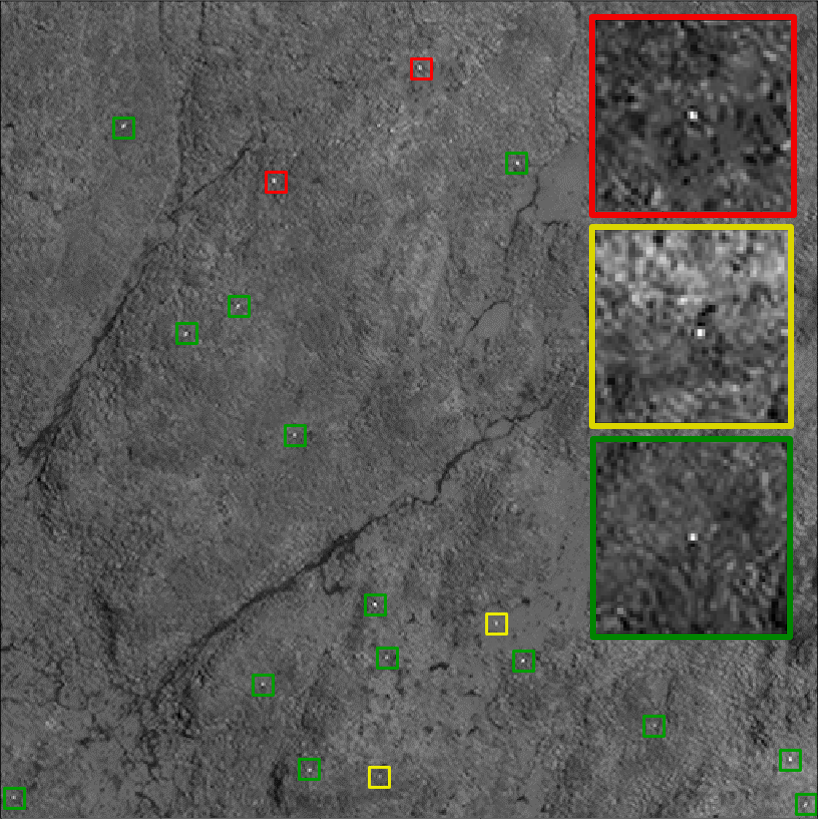}
\caption{Example of visually similar false positive (red), false negative (yellow) and true positive (green) detections.}
\label{fig:patchres}
\end{figure}

\section{EXPERIMENTAL RESULTS}
To quantitatively evaluate results, we sort detections into true positives (TPs), false positives (FPs) and false negatives (FNs) based on ground truth labels. Results are reported in terms of recall (the fraction of known albatrosses detected) and precision (the fraction of detections which are correct), calculated as:
\vspace{-3mm}
\begin{equation}
\text{Recall}=\dfrac{TPs}{TPs+FNs},\quad \text{Precision}=\dfrac{TPs}{TPs+FPs}
\end{equation}
We plot precision-recall curves by thresholding the model predictions at different probability values.

\subsection{Focal loss experiments}
\label{ssec:gamma}
We used the training set to generate three models for each choice of $\gamma$, and took the average of these three results on the test set (Fig \ref{fig:Gamma}). Our experiments showed that the network performed poorly and was very unstable for $\gamma=0$, indicating the need for a focusing parameter. For higher $\gamma$ values we found a strong improvement in the precision-recall curves. Although the difference between $\gamma=1$ and $\gamma=2$ was marginal, the training became more unstable for $\gamma=2$ (and higher), with the network quickly overfitting. With this in mind we select $\gamma=1$ as the best choice. 

With these parameters we achieve precision-recall values close to 80\% each on the test set (although adjusting the probability threshold can alter the precision-recall trade off). However on visual inspection, misclassifications are difficult to discern from apparently correct detections (Fig \ref{fig:patchres}). This highlights the problem with assessing model performance against ground truth data which has a high level of uncertainty.     

\begin{figure*}[t!]

\begin{minipage}[b]{0.45\linewidth}
\centering
\includegraphics[width=\linewidth]{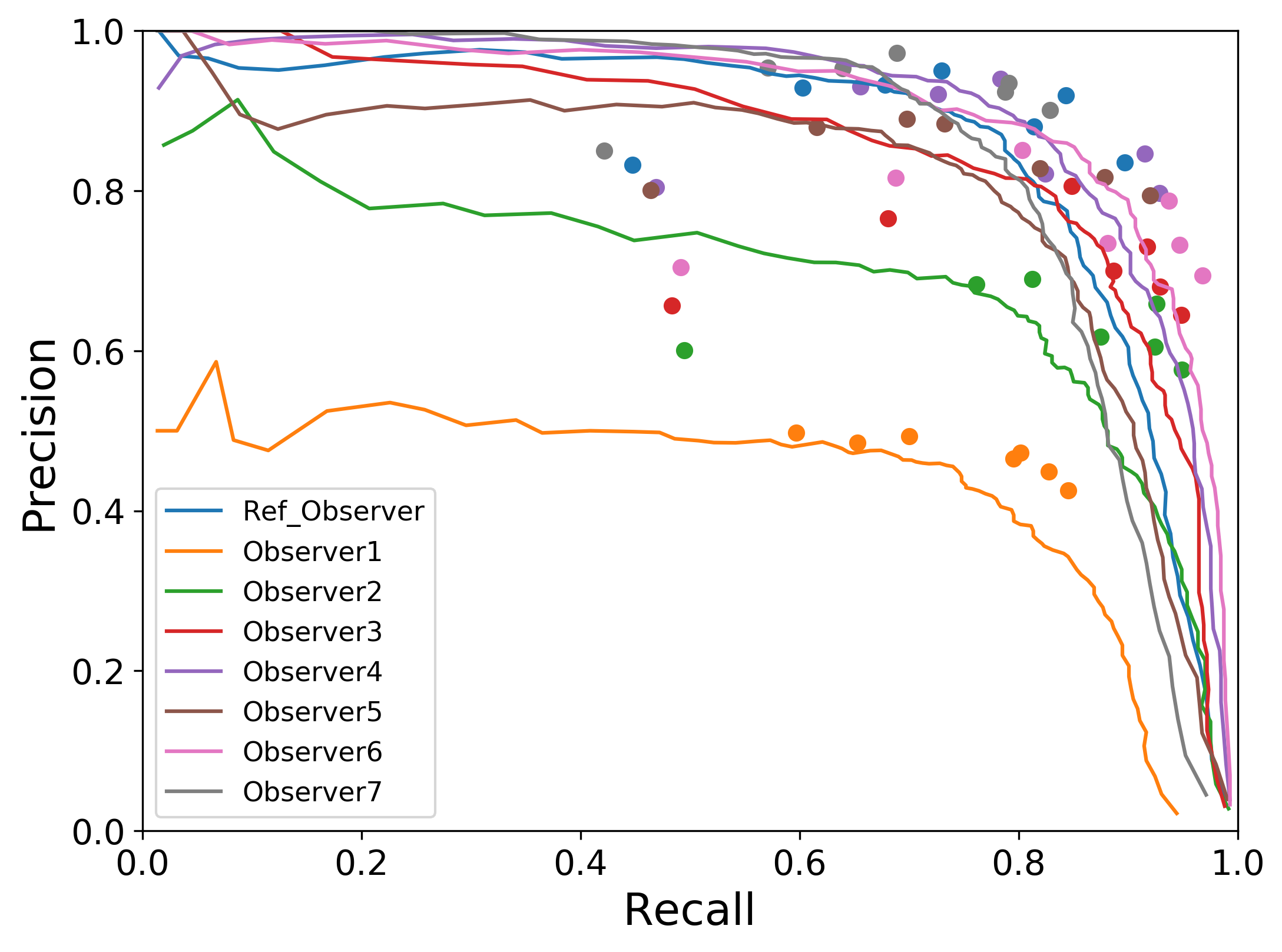}
\vspace{0.5cm}
\centerline{(a) Precision-recall curves for each observer}
\end{minipage} 
\begin{minipage}[b]{0.54\linewidth}
\centering
\includegraphics[width=\linewidth]{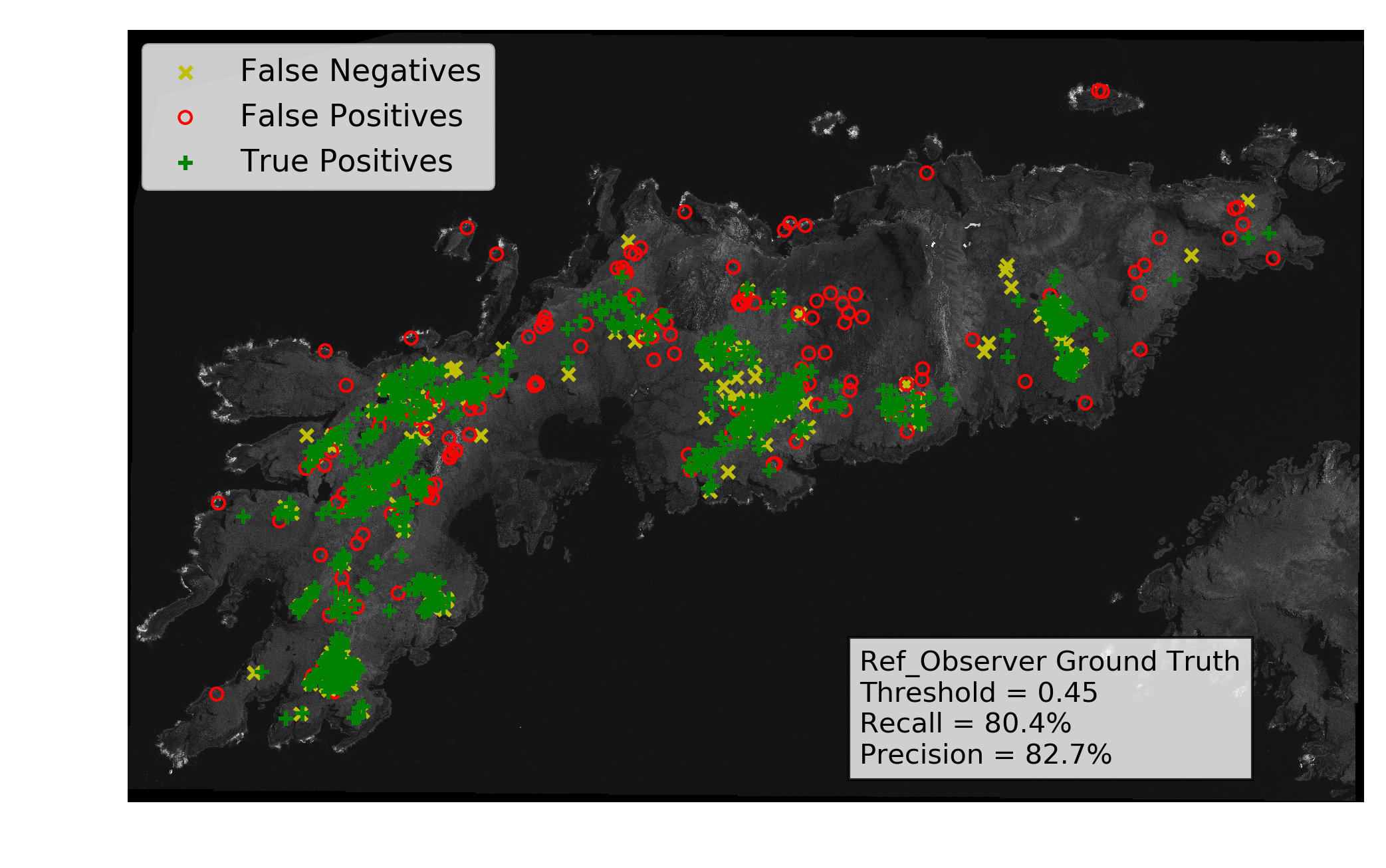}
\vspace{0.5cm}
\centerline{(b) Example detection results} 
\end{minipage} 
\vspace{-0.5cm}
\caption{Results of the model on the Bird Island dataset. (a) shows precision-recall results for the model (lines) and all other observers (points), using each of the eight observers' ground truth labels in turn. An example output (b) shows FPs, FNs and TPs corresponding to the reference observer's ground truth labels at probability threshold 0.45.}
\label{fig:InterObserver}
\end{figure*}

\subsection{Inter-observer variation assessment}
\label{ssec:interob}
The results of our inter-observer variation assessment are presented in Fig \ref{fig:InterObserver}. To compare results we take a single observers' labels as ground truth, and plot the precision-recall curve using the model predictions. These are shown as the lines in Fig \ref{fig:InterObserver}a. We additionally plot precision-recall points, showing how closely each other observer matched the chosen ground truth (points in Fig \ref{fig:InterObserver}a). This was repeated for all eight sets of ground truth points (note 'reference observer' refers to the analyst who labelled the three images used in the training set). 

Our results show how the ground truth labels can significantly influence the assessment of model performance. For example comparing results against the observer with the lowest count (observer 1 with 508), the precision is low. This is because the network detects many false positives, which are interpreted as albatrosses by most other observers. While this particular result could be viewed as an outlier, all observers show disagreement between labels (shown by the spread of points in Fig \ref{fig:InterObserver}a). Within this context we see that the model performs within the range of inter-observer variation, and therefore matches human performance. 

In Fig \ref{fig:InterObserver}b we show an example of detection results over Bird Island, using ground truth from the reference observer. We achieve precision of 82\% and recall of 80\% for the chosen probability threshold (0.45). At this operating point the network would output a prediction of 909 albatrosses (752 TPs + 157 FPs), in comparison to the 935 labelled by the observer. Future work will involve a more thorough assessment of inter-observer variation. Consideration will be given to the best practice for integrating uncertainty in the training phase (i.e through training using multiple observer's labels), as well as statistical methods for assessing final detections. For this we may draw inspiration from research in medical image analysis, where similar challenges arise.

\section{CONCLUSIONS}
\label{sec:conclusion}
In this paper we present an automated method for detecting Wandering Albatrosses in WV-3 satellite imagery. We show that a comparatively simple U-net architecture, in combination with the Focal Loss, can produce results in line with those achieved by human counters. We also highlight the significant level of uncertainty in ground truth labels, and stress the importance of building this into model development and assessment. We hope the methods will streamline and standardise satellite survey methods, for Wandering Albatrosses and potentially other species. 

\section{ACKNOWLEDGEMENTS}
\label{sec:acknowledgements}
This work was supported by NERC and EPSRC through the NEXUSS CDT [grant number NE/RO12156/1]. The WV-3 imagery used was courtesy of DigitalGlobe, a Maxar company. We gratefully acknowledge the support of NVIDIA Corporation with the donation of the Titan Xp GPU used for this research.


%
\bibliographystyle{IEEEbib}
\bibliography{refs}

\end{document}